\documentclass{article}
\usepackage{spconf,amsmath,graphicx}
\usepackage{url}
\newcommand\norm[1]{\left\lVert#1\right\rVert}
\usepackage{subcaption}
\usepackage{multirow,paralist}

\title{Learning ASR-Robust Contextualized Embeddings \\for Spoken Language Understanding}
\name{Chao-Wei Huang\qquad Yun-Nung Chen}
\address{National Taiwan University, Taipei, Taiwan\\
{\tt f07922069@csie.ntu.edu.tw\quad y.v.chen@ieee.org}}
\begin{document}
%
\maketitle
\begin{abstract}
  Employing pre-trained language models (LM) to extract contextualized word representations has achieved state-of-the-art performance on various NLP tasks.
  However, applying this technique to noisy transcripts generated by automatic speech recognizer (ASR) is concerned.
  Therefore, this paper focuses on making contextualized representations more ASR-robust.
  We propose a novel confusion-aware fine-tuning method to mitigate the impact of ASR errors on pre-trained LMs. Specifically, we fine-tune LMs to produce similar representations for acoustically confusable words that are obtained from word confusion networks (WCNs) produced by ASR. Experiments on multiple benchmark datasets show that the proposed method significantly improves the performance of spoken language understanding when performing on ASR transcripts\footnote{Code available at: \url{https://github.com/MiuLab/SpokenVec}}.
\end{abstract}


\section{Introduction}

  A spoken language understanding (SLU) module serves an important role in a spoken dialogue system, which aims at extracting semantic concepts from spoken utterances and provides structured information for accessing the backend database.
  Typical tasks of SLU include intent detection and slot filling. 
  These two tasks focus on predicting speaker's intent and extracting semantic concepts as constraints for the natural language. 
  A movie-related example utterance ``{\em find comedies by James Cameron}'' shown in Figure~\ref{fig:iob} has two slot-value labels and a specific intent for the whole utterance.
  
  Applying deep learning techniques has been shown to boost the performance of SLU \cite{yao2014spoken,guo2014joint,mesnil2014using,goo-etal-2018-slot}.
  Most prior work focused on applying understanding models on \emph{manual} transcripts, ignoring the errors introduced by automatic speech recognizers (ASR).
  Hence, several methods were proposed to address this problem. Simonnet et al.~\cite{simonnet-etal-2018-simulating} simulated ASR errors and trained SLU models for better handling of the errors.  
  The prior work leveraged information from lattices or word confusion networks~\cite{hakkani2006beyond,Tr2013SemanticPU,ladhak2016latticernn,shivakumar2018confusion2vec,shivakumar2019spoken,huang2019adapting},
  and Zhu et al.~\cite{zhu2018robust} applied domain adversarial training for ASR-error adaptation, demonstrating the importance of incorporating ASR errors for better SLU performance.

\begin{figure}[t]
\centering
\begin{tabular}{|cl|}
\hline
{\bf Word} & \emph{find comedies by james cameron}\\
{\bf Slot} & genre: \emph{comedy}, director: \emph{James Cameron} \\
{\bf Intent} & find\_movie \\
\hline
\end{tabular}
\caption{An annotated utterance example.}
\label{fig:iob}
\end{figure}

Deep contextualized word representations recently have achieved great success among language understanding tasks \cite{peters2018deep,devlin2018bert,siddhant2018unsupervised}.
Nevertheless, they may be less robust to noisy texts, such as the recognized results.
In this paper, we investigate the impact of ASR errors on contextualized embeddings and further propose a novel confusion-aware fine-tuning method to alleviate this problem.
To our best knowledge, there is no prior work that learned contextualized word embeddings and considered the errors produced from spoken language for better robustness.
Our contributions are 3-fold:
\begin{itemize}
  \item This is the first attempt to learn contextualized word embeddings specifically for spoken language.
  \item The proposed approach achieves better performance on the benchmark spoken language understanding tasks.
  \item The proposed method shows better robustness to ASR errors.
\end{itemize}

\section{Learning ASR-Robust Contextualized Embeddings}

Preparing datasets for SLU takes a lot of effort. SLU datasets are typically smaller than NLU datasets since it's much more labor intensive to collect labeled spoken utterances. Hence, the goal of this paper is to learn ASR-robust contextualized embeddings such that the downstream SLU models trained on manual transcripts can perform well on automatic transcripts, using only unlabeled spoken utterances to adapt.

To enable the embeddings to adapt ASR errors for improving SLU, our proposed method consists of three stages:
1) language model pre-training on general domain corpora $\mathcal{D}_\text{LM}$,
2) confusion-aware language model fine-tuning on the text from the target SLU task, where the text can be either manual transcripts $\mathcal{D}_\text{trs}$ or automatic transcripts $\mathcal{D}_\text{asr}$, and
3) training a language understanding model with the fine-tuned LM on labeled SLU data $\mathcal{D}_\text{SLU}$, which consists of the manual transcripts $\mathcal{D}_\text{trs}$ with their corresponding labels.

In this paper, we focus on the task of intent detection, which is an utterance-level multi-class classification problem.
More formally, given an utterance $x = \{w_{1}^{x}, w_{2}^{x}, ..., w_{|x|}^{x}\}$, the goal is to predict its corresponding intent $I_{x}$.
The input utterance $x$ can be either manually transcribed texts, denoted as $x_\text{trs}$, or ASR-recognized results, denoted as $x_\text{asr}$.
The proposed approach is detailed below.
  
\subsection{Embeddings from Language Model (ELMo)}
Peters et al.~\cite{peters2018deep} proposed ELMo to extract context-dependent word embeddings from a pre-trained LM, and the contextualized embeddings were proved to be able to improve the performance of downstream NLP tasks.
In this paper, we adopt the same model architecture as in the original work, which consists of a CNN character encoder and two bidirectional LSTMs \cite{hochreiter1997long}.
Same strategy of combining hidden states from different layers is applied~\cite{peters2018deep}, which computes the representation $e_t$ for a word $w_{t}^{x}$ in the sentence $x$ as:
\begin{equation*}
    e_t = \gamma \sum_{i=0}^{2} {\alpha_{i} \cdot h_{t,i}^{x}},
\end{equation*}
where $h_{t,i}^{x} = [ \overleftarrow{h_{t,i}^{x}}; \overrightarrow{h_{t,i}^{x}} ]$ is the concatenation of the $i$-th layer output from both directions at the time $t$, $\alpha_i$ is the weight for the $i$-th layer, and $\gamma$ is a scaling factor. $\alpha_i$ and $\gamma$ are scalar parameters learned along with downstream tasks.
The ELMo model is pre-trained on the general-domain textual data $\mathcal{D}_\text{LM}$.

\subsection{Language Model Fine-Tuning}
One advantage of pre-training a language model is that it can leverage large amounts of unlabeled text corpora.
Usually the data is general such as Wikipedia.
However, the data distribution of the target task may be different from that used in pre-training, posing a domain mismatch problem.
Howard et al.~\cite{howard-ruder-2018-universal} proposed to fine-tune the pre-trained LM with sentences from the downstream dataset and showed that it boosts the performance of the downstream task. Chronopoulou et al.~\cite{chronopoulou2019embarrassingly} also demonstrated the effectiveness of the fine-tuning method.

In order to adapt the pre-trained LM to the target data, the fine-tuning technique is applied.
Given an utterance $x = \{w_1, w_2, ..., w_{|x|}\}$, the bidirectional language modeling loss can be written as:
\begin{equation*}
    \mathcal{L}_\text{LM} = \frac{1}{|x|} \sum_{t=1}^{|x|} - \log p(w_t \mid w_{<t}) - \log p(w_t \mid w_{>t}),
\end{equation*}
where $p(w_t \mid w_{<t})$ and $p(w_t \mid w_{>t})$ are probabilities of $w_t$ predicted by the forward LM and the backward LM respectively.

Language model fine-tuning can be performed on both manual transcription and recognized results

\begin{figure*}[t!]
\centering
\captionsetup[subfigure]{justification=centering}
\begin{subfigure}[t]{0.44\textwidth}
    \centering
    \includegraphics[width=\linewidth]{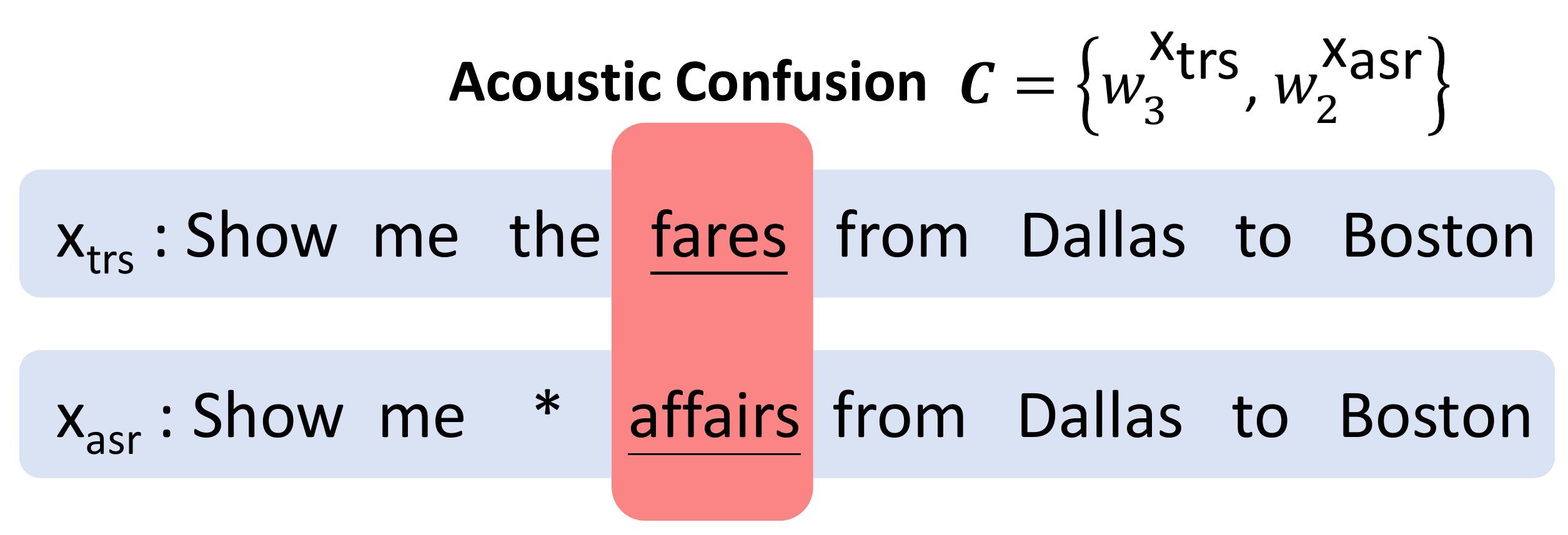}
    \caption{Supervised confusion extraction \\ with aligned utterances}
    \label{fig:sup-conf}
\end{subfigure}
~
\hfill
\begin{subfigure}[t]{0.54\textwidth}
    \centering
    \includegraphics[width=\linewidth]{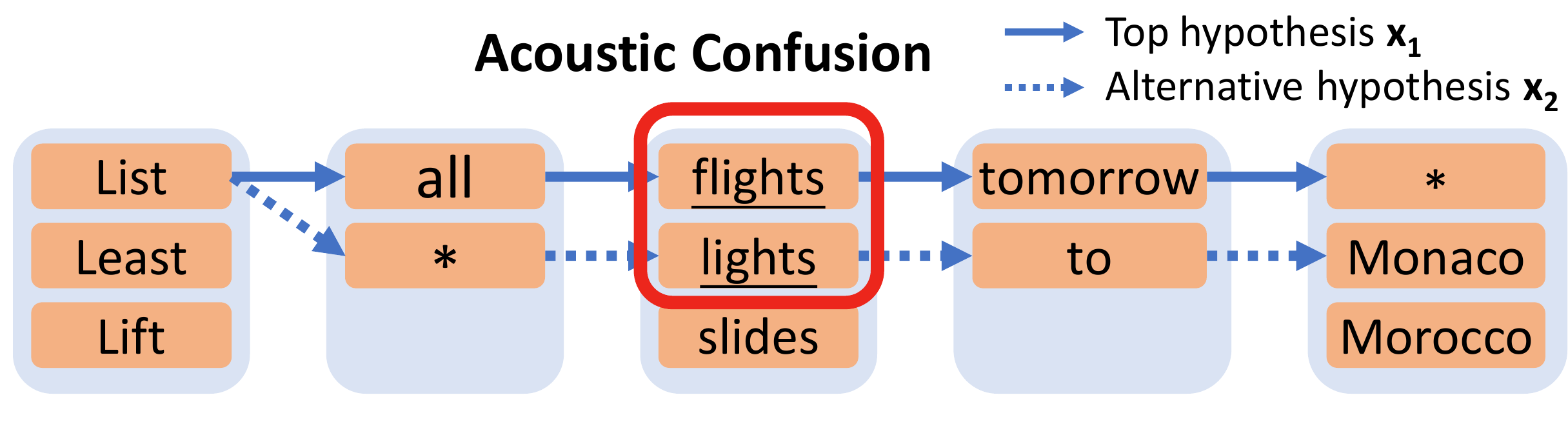}
    \caption{Unsupervised confusion extraction using \\ a WCN generated by ASR}
    \label{fig:unsup-conf}
\end{subfigure}
\vspace{-1mm}
\caption{Illustration of different extraction approaches. $*$ denotes a blank symbol for alignment purpose.}
\label{fig:conf}
\vspace{-2mm}
\end{figure*}

\subsection{Confusion-Aware Fine-Tuning}
Taking ASR transcripts as inputs may introduce an issue that words in an utterance may be misrecognized.
For instance, {\em fair} and {\em fare} are acoustically similar, so an ASR system may fail to distinguish between them, resulting in a substitution error. 
Such recognition errors might be recovered by human, because human are aware of the acoustic confusability of words.
However, the errors may significantly degrade the testing performance when the models are trained on manual transcripts.
In order to enhance the ASR robustness in contextualized word embeddings, this section integrates the acoustic confusion into our LM.

We propose a confusion-aware fine-tuning method to mitigate this problem from pre-trained LMs, which aims at making the LM consider multiple acoustically confusable words. 
Let $c = \{w_{t_1}^{x_1}, w_{t_2}^{x_2}\}$ denote an acoustic confusion, i.e., two words with similar pronunciation in two different utterances $x_1$ and $x_2$, and $C = \{c_1, c_2, \cdots, c_{|C|}\}$ denote the set of all acoustic confusions in $x_1$ and $x_2$.
We introduce a new loss term called confusion loss:
\begin{equation*}
    \mathcal{L}_\text{conf} = \frac{1}{|C|} \sum_{c \in C} \sum_{i=0}^{1} {1 - \frac{h_{t_1,i}^{x_1} \cdot h_{t_2,i}^{x_2}}{\norm{h_{t_1,i}^{x_1}} \norm{h_{t_2,i}^{x_2}}}},
\end{equation*}
which is the cosine distance between the LM hidden states corresponding to words. 
Note that we empirically find that including only the first two layers in loss computation works the best.
Two approaches are designed for extracting acoustic confusions.

\subsubsection{Supervised Confusion Extraction}
Assuming that both ASR transcripts $x_\text{asr}$ and manual transcripts $x_\text{trs}$ of a spoken utterance are accessible, we align $x_\text{asr}$ with $x_\text{trs}$ with respect to minimum edit-distance criterion to extract acoustic confusions as shown in Figure~\ref{fig:sup-conf}.
By minimizing $\mathcal{L}_\text{conf}$, we directly force the LM to produce representations for an erroneous word similar to its correct counterpart.
This method is called \emph{supervised confusion extraction} considering that it requires the availability of manual transcripts $\mathcal{D}_\text{trs}$.

\subsubsection{Unsupervised Confusion Extraction}
Considering the scenario where only audio recording of a spoken utterance is available, we can apply an ASR on the recording and construct a word confusion network (WCN).
Then a list of n-best hypotheses is generated and aligned using WCN, and the acoustic confusions can be obtained as depicted in Figure~\ref{fig:unsup-conf}.

An important advantage of this approach is that it does not require any labeled utterances or manual transcripts; therefore, we can leverage unlabeled audio recordings to fine-tune LMs in an unsupervised fashion.

\subsection{Joint Objective Function for Fine-tuning}
In the fine-tuning stage, we minimize the joint objective function including the LM loss and confusion-aware loss:
\begin{equation*}
    \mathcal{L}_\text{FT} = \mathcal{L}_\text{LM} + \beta \mathcal{L}_\text{conf},
\end{equation*}
where $\beta$ is a hyperparameter to balance the contribution of two loss functions.
The procedure enables our model to incorporate not only the target domain information but the acoustic information for better robustness to ASR errors.

\begin{figure}[t!]
    \includegraphics[width=0.9\linewidth]{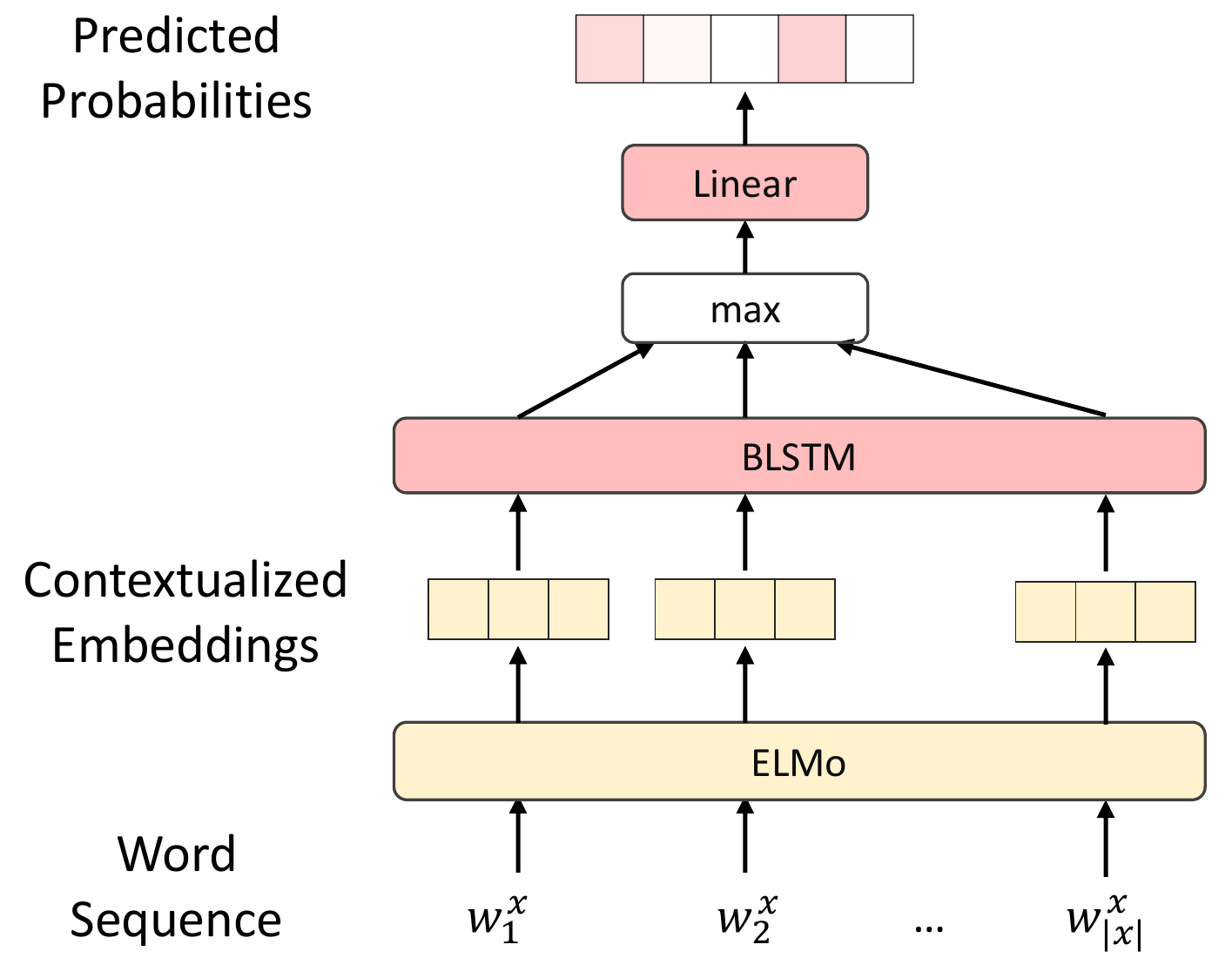}
    \vspace{-1mm}
    \caption{Illustration of our SLU model architecture.}
    \label{fig:arch}
    \vspace{-3mm}
\end{figure}

\subsection{Spoken Language Understanding (SLU)}
To further build an SLU model that leverages ASR-robust contextualized embeddings, we employ a biLSTM as our SLU model, where the biLSTM takes contextualized word embeddings $\{e_t\}_{t=1}^{|x|}$ as the input, and the outputs of the last biLSTM layer are max-pooled, linearly transformed and softmaxed to obtain the predicted probabilities for each class.
The overall architecture is illustrated in Figure~\ref{fig:arch}. During training, we use cross entropy as the loss function.
Weights of the ELMo model are fixed during this stage except for $\alpha_i$ and $\gamma$.
The SLU model is trained on $\mathcal{D}_\text{SLU}$ and evaluated on automatic transcripts. The trained SLU model is expected to achieve better performance on automatic transcripts due to the integration of ASR-robust contextualized word embeddings.

\section{Experiments}

\begin{table}[h]
\centering
\begin{tabular}{|l|c||c|c|c|c|}
\hline
\multicolumn{2}{|l||}{}                 & \textbf{train}       & \textbf{test}       & \textbf{intents} & \textbf{WER}     \\
\hline\hline
\multicolumn{2}{|c||}{ATIS}             & 4478        & 893        & 21         & 15.55\% \\
\hline
\multirow{2}{*}{SmartLights} & close & \multicolumn{2}{c|}{1765} & 6          & 45.61\% \\
\cline{2-6}
                             & far   & \multicolumn{2}{c|}{1765} & 6          & 71.02\% \\
\hline
\multicolumn{2}{|c||}{Snips}            & 13084       & 700        & 7          & 45.56\% \\
\hline
\end{tabular}
\caption{Dataset statistics.}
\label{tab:dataset}
\vspace{-3mm}
\end{table}

\begin{table*}[t]
\centering
\begin{tabular}{|c|l|c|c|c|c|c|c|c|c|}
\hline
    \multicolumn{2}{|c|}{\multirow{3}{*}{\bf Model}} & \multicolumn{2}{c|}{\multirow{2}{*}{\bf ATIS}} & \multicolumn{4}{c|}{\bf SmartLights} & \multicolumn{2}{c|}{\multirow{2}{*}{\bf Snips}} \\
\cline{5-8}
    \multicolumn{2}{|c|}{} & \multicolumn{2}{c|}{} & \multicolumn{2}{c|}{\bf close} & \multicolumn{2}{c|}{\bf far} & \multicolumn{2}{c|}{} \\
\cline{3-10}
    \multicolumn{2}{|c|}{} & \multicolumn{1}{c|}{\bf Manual} & \multicolumn{1}{c|}{\bf ASR} & \multicolumn{1}{c|}{\bf Manual} & \multicolumn{1}{c|}{\bf ASR} & \multicolumn{1}{c|}{\bf Manual} & \multicolumn{1}{c|}{\bf ASR} & \multicolumn{1}{c|}{\bf Manual} & \multicolumn{1}{c|}{\bf ASR} \\
\hline\hline
(a) & Oracle            & 96.47 & 94.75 & 90.60 & 74.05 & 82.94 & 56.96 & 96.38 & 91.79 \\
\hline
(b) & Context-independent          & 93.60 & 90.35 & 95.67 & 65.71 & 95.67 & 44.55 & 96.29 & 72.70 \\
(c) & Pre-trained ELMo             & 96.65 & 93.27 & 97.01 & 64.53 & 97.01 & 44.46 & 96.29 & 77.86 \\
\hline
(d) & (c) + fine-tune, $\mathcal{L}_\text{LM}$ only & 96.91 & 94.27 & 95.91 & 66.33 & 95.66 & 46.22 & 96.38 & 87.74 \\
(e) & (c) + fine-tune, $\mathcal{L}_\text{FT}$ (sup-conf) & 96.61 & \bf 95.65 & 95.92 & \bf 67.99 & 95.53 & 46.57 & 97.01 & 88.52 \\
(f) & (c) + fine-tune, $\mathcal{L}_\text{FT}$ (unsup-conf) & 97.02 & 95.39 & 95.98 & 67.98 & 95.79 & \bf 47.38 & 97.04 & \bf 89.55\\
\hline
\end{tabular}
\vspace{-1mm}
\caption{Results of intent detection tasks (\%). \textbf{Manual} and \textbf{ASR} indicate evaluating on $x_\text{trs}$ and $x_\text{asr}$ respectively. \textbf{close} and \textbf{far} represent different microphone settings. \textit{sup-conf} stands for supervised confusion extraction, and \textit{unsup-conf} stands for unsupervised confusion extraction. The best numbers for each dataset are marked in bold.} 
\label{tab:result}
\vspace{-3mm}
\end{table*}

\subsection{Setup}
Three SLU datasets used in the experiments as listed below:
\begin{compactitem}
    \item ATIS (Airline Travel Information Systems) ~\cite{hemphill1990atis,dahl1994expanding,tur2010left} is a benchmark dataset widely used in language understanding research.
    The dataset contains audio recordings of people making flight reservations with corresponding manual transcripts. 
    \item Snips SmartLights \cite{saade2018spoken} contains spoken commands for smart light assistant. The dataset comes with two kinds of microphone settings, \textit{close field} and \textit{far field}.
    \item Snips \cite{coucke2018snips} is a dataset for benchmarking NLU systems. This dataset is larger than ATIS and Snips SmartLights. We use a commercial text-to-speech system \footnote{https://cloud.google.com/text-to-speech/} to synthesize audio from text data.
\end{compactitem}
The dataset statistics are shown in Table \ref{tab:dataset}.

For ATIS, we train an ASR system on WSJ~\cite{Paul:1992:DWS:1075527.1075614} using the \textit{s5} recipe from Kaldi \cite{povey2011kaldi}. For the other datasets, we use an ASR model released in Kaldi to provide better recognition results \footnote{https://kaldi-asr.org/models/m1}.
We use the ASR system to recognize audio recordings and extract acoustic confusions for fine-tuning.

\subsection{Model and Training Details}
The pre-trained weights of ELMo from \cite{peters2018deep} 
are adopted.
The size of contextualized representations is 1024. 
Our SLU model has two layers with 300-dimensional hidden states.

In the fine-tuning stage, acoustic confusions that contain stop words are excluded, and $\beta$ is set to 0.1. We set batch size to 64 and use \texttt{Adam} as the optimizer \cite{kingma2014adam} with learning rate 0.001 for all stages.
We fine-tune ELMo for 3 epochs and train the SLU model for 50 epochs.
The Snips SmartLights dataset is very small, so we use 10-fold cross validation to evaluate the models as suggested in \cite{saade2018spoken}.

\subsection{Baselines}
We compare our method with two baselines and an oracle system as listed below.
\begin{compactitem}
    \item Context-independent: replaces the contextualized representations with traditional context-independent word embeddings. The embedding matrix is initialized randomly.
    \item Pre-trained ELMo: uses pre-trained ELMo weights without fine-tuning.
    \item $\text{Oracle}$: trains SLU on $x_\text{asr}$ with pre-trained ELMo embeddings.
\end{compactitem}
Note that our models do not utilize the information the oracle system uses, $x_\text{asr}$ paired with labels. 

\subsection{Results}
Table \ref{tab:result} shows the experimental results, where the reported numbers are accuracies averaged over 5 runs. 
All models are trained on $x_\text{trs}$ except for the oracle system, and they all perform great when evaluated on $x_\text{trs}$.
Rows (b) and (c) show that ASR errors degrade SLU performance considerably for both context-independent and context-dependent embeddings. When testing on $x_\text{asr}$, the performance drops 3.38\% on ATIS dataset using pre-trained ELMo embeddings. The results on Snips datasets show that the performance drops more in higher WER scenarios.

Our proposed method, confusion-aware language model fine-tuning, outperforms baselines by a large margin on all datasets (rows (d)-(f)), while it maintains identical performance on $x_\text{trs}$.
Results in row (d) can be viewed as an ablation to rows (e) and (f), where we exclude $\mathcal{L}_\text{conf}$ from the joint objective. 
Rows (e) and (f) show that while $\mathcal{L}_\text{LM}$ provides significant improvement alone, adding $\mathcal{L}_\text{conf}$ further boosts performance notably.
The results demonstrate that the proposed method can provide ASR robustness to the SLU models. 

\subsection{Discussion}
The research on SLU has been investigated for several years, and there are two main branches.
The first branch treats SLU as a natural language understanding (NLU) task, where the prior work trained the models directly on natural language data without misrecognition~\cite{yao2014spoken,guo2014joint,mesnil2014using,goo-etal-2018-slot}.
Another branch focuses on building understanding models with consideration of ASR results. While some prior work relied on ASR lattices or WCNs to provide richer information to SLU models~\cite{hakkani2006beyond,Tr2013SemanticPU,ladhak2016latticernn}, the work presented here focuses on using the 1-best results from ASR.
Simonnet et al. proposed an ASR error simulation scheme to train robust SLU models~\cite{simonnet-etal-2018-simulating}, whereas we dig realistic ASR errors from recognition results. Shivakumar et al. used WCNs to extract acoustic confusions for fine-tuning context-independent word embeddings~\cite{shivakumar2018confusion2vec,shivakumar2019spoken}. Our work combines the idea of using acoustic confusions with language model fine-tuning~\cite{howard-ruder-2018-universal} to obtain ASR-robust contextualized embeddings.

\section{Conclusion}
This paper proposes a novel confusion-aware language model fine-tuning method for learning ASR-robust contextualized embeddings.
We introduce supervised and unsupervised methods for extracting acoustic confusions and integrate a confusion loss that forces LMs to consider acoustically confusable words. The experiments on SLU demonstrate that our proposed method learns contextualized embeddings that are robust to ASR errors.

\bibliographystyle{IEEEbib}
\bibliography{strings,refs}

\end{document}